\pdfoutput=1
\documentclass{article}
\usepackage{ijcai25}

\usepackage{times}
\usepackage{soul}
\usepackage{url}
\usepackage[hidelinks]{hyperref}
\usepackage[utf8]{inputenc}
\usepackage[small]{caption}
\usepackage{graphicx}
\usepackage{amsmath}
\usepackage{amsthm}
\usepackage{booktabs}
\usepackage{algorithm}
\usepackage{algorithmic}
\usepackage{CJKutf8}
 \usepackage{multirow}
\usepackage{graphicx} 
\usepackage[switch]{lineno}

\usepackage{latexsym}

\title{In-Context Meta LoRA Generation}

\author{
Yihua Shao*$^1$
\and
Minxi Yan*$^2$
\and
Yang Liu$^3$
\and
Siyu Chen$^2$
\and
Wenjie Chen$^2$
\and
Xinwei Long$^2$
\and
\\Ziyang Yan$^4$
\and
Lei Li$^5$
\and
Chenyu Zhang$^4$
\and
Nicu Sebe$^4$
\and
Hao Tang$^6$
\and
Yan Wang$^2$
\and
Hao Zhao$^2$
\and
\\Mengzhu Wang$^1$
\and
Jingcai Guo$^1$
\\
\affiliations
$^1$The Hong Kong Polytechnic University\\
$^2$Tsinghua University\\
$^3$Beijing Institute for General Artificial Intelligence (BIGAI)\\
$^4$University of Trento\\
$^5$University of Copenhagen\\
$^6$Peking University\\
\emails
haotang@pku.edu.cn, jc-jingcai.guo@polyu.edu.hk
}

\begin{document}
\begin{CJK}{UTF8}{gbsn}
\maketitle

\begin{abstract}
Low-rank Adaptation (LoRA) has demonstrated remarkable capabilities for task specific fine-tuning. However, in scenarios that involve multiple tasks, training a separate LoRA model for each one results in considerable inefficiency in terms of storage and inference. Moreover, existing parameter generation methods fail to capture the correlations among these tasks, making multi-task LoRA parameter generation challenging. To address these limitations, we propose In-Context Meta LoRA (ICM-LoRA), a novel approach that efficiently achieves task-specific customization of large language models (LLMs). Specifically, we use training data from all tasks to train a tailored generator, Conditional Variational Autoencoder (CVAE). CVAE takes task descriptions as inputs and produces task-aware LoRA weights as outputs. These LoRA weights are then merged with LLMs to create task-specialized models without the need for additional fine-tuning. Furthermore, we utilize in-context meta-learning for knowledge enhancement and task mapping, to capture the relationship between tasks and parameter distributions. As a result, our method achieves more accurate LoRA parameter generation for diverse tasks using CVAE. ICM-LoRA enables more accurate LoRA parameter reconstruction than current parameter reconstruction methods and is useful for implementing task-specific enhancements of LoRA parameters. At the same time, our method occupies 283MB, only 1\% storage compared with the original LoRA.
\end{abstract}

\section{Introduction}

Large-scale models (LLMs/MLMs) have become the cornerstone of modern AI applications \cite{achiam2023gpt,xiao2024florence,dubey2024llama,shao2024gwq}. However, these models typically require substantial amounts of data for fine-tuning. We always fine-tune LLMs with Low-rank Adaptation (LoRA) \cite{hu2021lora} using task-specific data. In scenarios with numerous sub-tasks \cite{erkocc2023hyperdiffusion,yan20243dsceneeditor}, the current approach of training a separate LoRA for each sub-task leads to inefficiency in storage and inference. For instance, in multi-task scenarios, the weights of LoRA can become prohibitively expensive to store, necessitating more efficient solutions. Although FM-Delta~\cite{mizrahi2017fm} employs a novel compression scheme that significantly reduces the need for storage by storing compressed fine-tuned models, it does not address the issue of capturing correlations between sub-tasks. 

The current parameter generation methods \cite{platanios2018contextual,wortsman2022model,jin2024conditional} can only implement the generation of LoRA parameters for a single task, and it is not possible to implement the simultaneous generation of LoRA weights required for different tasks through only one generator. Moreover, the current parameter generation training method lacks context modeling capability, which makes it difficult to implement multi-use multi-task enhancement of LoRA weights. This causes a great storage burden when storing LoRA weights and training data.
\begin{figure}
    \centering
    \includegraphics[width=0.8\linewidth]{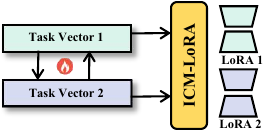}
    \caption{ICM-LoRA achieves accurate reconstruction of LoRA parameters through task vectors context modeling.}
    \label{fig:pip}
\end{figure}

Therefore, we propose the \textbf{In-Context Meta LoRA (ICM-LoRA)} to implement generating different tasks LoRA with self-designed generator, Conditional Variational Autoencoder (CVAE). As shown in figure \ref{fig:pip}, ICM-LoRA utilizes task vectors for context modeling through in-context meta learning, allowing the augmented CVAE to learn the parameter distribution features. By combining in-context learning and meta learning and using task vectors as modeling labels, we achieve meta enhancement for LoRA parameters. Furthermore, ICM-LoRA can get rid of the dependence on data and storage and only need to use a generator to implement the parameter generation.  

We evaluate our method on both text and visual tasks on different models. For visual tasks, we select target detection tasks and use the COCO dataset \cite{lin2014microsoft} to classify subclasses based on the detection task labels for experiments. For language tasks, we choose The Pile \cite{gao2020pile} as the training corpus and use five different subsets of it to simulate multi-class training tasks and validate the model on the validation set. The results indicate that CVAE could generate different tasks' LoRA parameters successfully. Compared to current methods, the generated LoRA parameters gain less accuracy loss. In addition, compared to original datasets and LoRA weights, our generator significantly reduce the storage.

In summary, the contributions of our approach can be summarized as follows:

\begin{itemize}
  \item [1)] 
    % We present a framework of a transformer-based accident warning agent: AccidentBlip, by vision-only inference to detect and predict the accident surround ego.
    We propose a novel framework, In-Context Meta LoRA (ICM-LoRA), which uses a self-designed parameter generator, Conditional Variational Autoencoder (CVAE), to generate LoRA weights, addressing the inefficiency of training separate LoRA models for multiple sub-tasks. 
  \item [2)]
    % We propose MA-former which could process temporal multi-view images by replace Q-former's Self-Attention to temporal attention.
    We employ in-context meta-learning for knowledge enhancement and task matching, which enables the generator to better learn the correspondence between tasks and model parameter distributions.
    \item [3)] 
    % AccidentBlip achieves recall precision and average precision on accident detection task and accident prediction accuracy on accident prediction task on DeepAccident dataset.
    Compared to existing methods, CVAE can generate task-specific LoRA parameters as same as the original LoRA or even better than the original LoRA. Also, our ICM-LoRA could cost only 1\% storage compared with original datasets.
\end{itemize}
\section{Related Works}
\subsection{Parameters Generation}
The core of parameter generation targets to help model generate similar distribution with original model. As one of the pioneers, \cite{platanios2018contextual} introduced a contextual parameter generator (CPG) addressing the challenge of training separate models for each language pair in neural machine translation (NMT). Some methods like stochastic neural networks \cite{sompolinsky1988chaos,bottou1991stochastic,wong1991stochastic,schmidt1992feed,murata1994network,graves2011practical} and Bayesian neural networks \cite{neal2012bayesian,kingma2013auto,rezende2014stochastic,kingma2015variational,blundell2015weight,gal2016dropout} improved the robustness and generalisation of the model through the prior probability distribution of the parameters, but these methods performed poorly in large-scale or complex scenarios. HyperNetworks~\cite{ha2016hypernetworks} generates parameters of large networks through small networks. With the development of diffusion, methods such as G.pt \cite{peebles2022learning} and P-diff families \cite{hu2021p,zhao2021p} began to use diffusion to generate normal scale parameters, but they are limited in generating parameters too large or too small. Furthermore, COND P-DIFF \cite{jin2024conditional} first applies parameter generation to generate LoRA parameters, but it only generates Lora models for coarse-grained tasks, and its parameters for generating LoRA for fine-grained tasks do not perform well. Therefore, we design a fine-grained task Lora generator which use in-context learning (Sec. \ref{ICL}) to enhance the ability of context understanding of generator model such as diffusion.

\subsection{In-Context Learning}
\label{ICL}
In-Context Learning (ICL) has emerged as a powerful paradigm in machine learning. As a pioneering work, \cite{brown2020language} reveals for the first time the learning ability of large language models in the presence of a small number of examples. Building upon this, for training LLMs, MetaICL \cite{min2021metaicl} integrated tasks into ICL format and enabled models to reach performance similar to direct fine-tuning. Lamda \cite{thoppilan2022lamda} emphasizes instruction tuning for models to understand better task descriptions instead of just examples. Self-instruct \cite{wang2022self} lets LLMs generate instructions for task alignment to explore enhancing ICL, while \cite{wei2022chain} introduced Chain-of-Thoughts (CoT) as an intermediate step between input and output to boost LLM reasoning in ICL. Task seperation ICL like Self-Ask \cite{press2022measuring} and ICAP \cite{chi2014icap} have explored multi-stage ICL, where tasks are broken down into simpler sub-tasks, each with its own set of demonstrations, and LLMs can process them individually. SuperICL \cite{xu2023small} utilizes smaller models as plugins to effectively execute tasks within the LLM framework, demonstrating the potential for hybrid model approaches in ICL. In scenario understanding, In-Context LoRA (IC-LoRA) \cite{huang2024context} and \cite{hendel2023context} respectively apply ICL on image generation by diffusion and context classification by LLM. In our paper, we apply ICL to generator to help generator better understand the context information in Lora.
\subsection{Dataset Condensation}
Dataset Condensation (DC) aims to create a compact and representative subset of the original training data. As a foundational work, \cite{zhao2020dataset} first tried to compress the data by matching the gradients of the synthetic dataset with the original, to ensure the condensed dataset retains the essential characteristics for effective model training. \cite{zhao2021dataset} further achieves more efficient data enhancement to synthesise more informative synthetic images by using differentiable twin enhancement (DSA). Building on it, \cite{zhao2023dataset} also explores DC with Distribution Matching, optimizing synthetic data to match the original distribution in embedding spaces via maximum mean discrepancy (MMD). \cite{wei2024dataset} compresses data by matching latent space quantiles and minimizing distribution fit statistics and \cite{lee2022dataset} further advances the field by modifying the loss function to capture class differences with a bi-level warm-up strategy for stable optimization. \cite{he2024multisize} integrates multiple dataset compression processes to get datasets of various sizes and introduced adaptive subset loss to reduce subset degradation. \cite{wang2022cafe} proposes a new method for dataset compression by aligning features, while \cite{liu2024dataset} introduces a dual-domain matching method for dataset condensation in time series classification, which further extends the applicability of DC to different data types. In our work, we discard the original dataset and deposit the different task information of the dataset into the generator model for data information aggregation and compression.

\begin{figure}[t] \small
    \centering
    \includegraphics[width=0.7\linewidth]{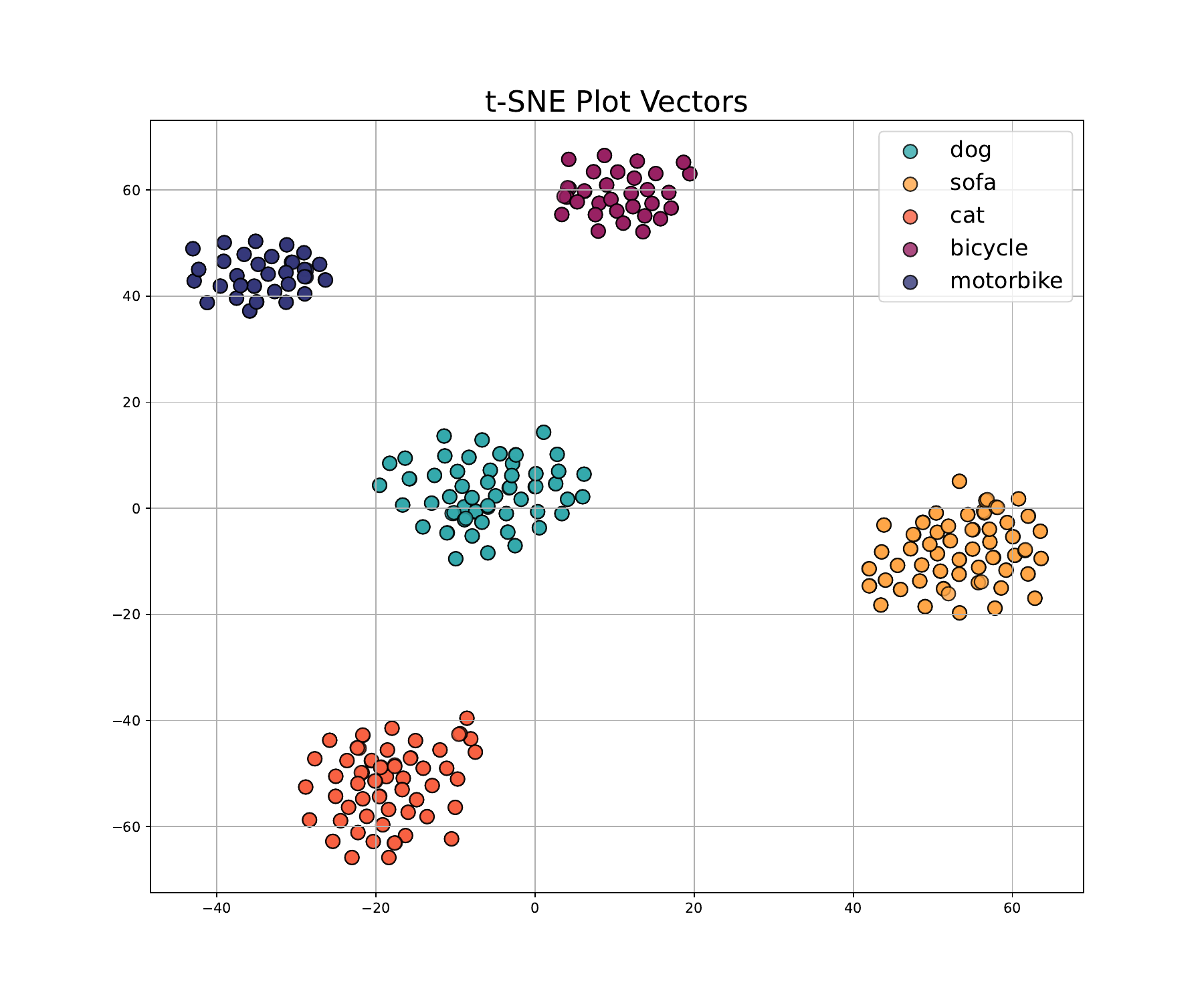}
    \caption{\textbf{Task Hidden Space Distribution. }Hidden states of tasks LoRA parameters have clustering phenomena. }
    \label{fig:sne} 
\end{figure}

\begin{figure*} \small
    \centering
    \includegraphics[width=0.85\linewidth]{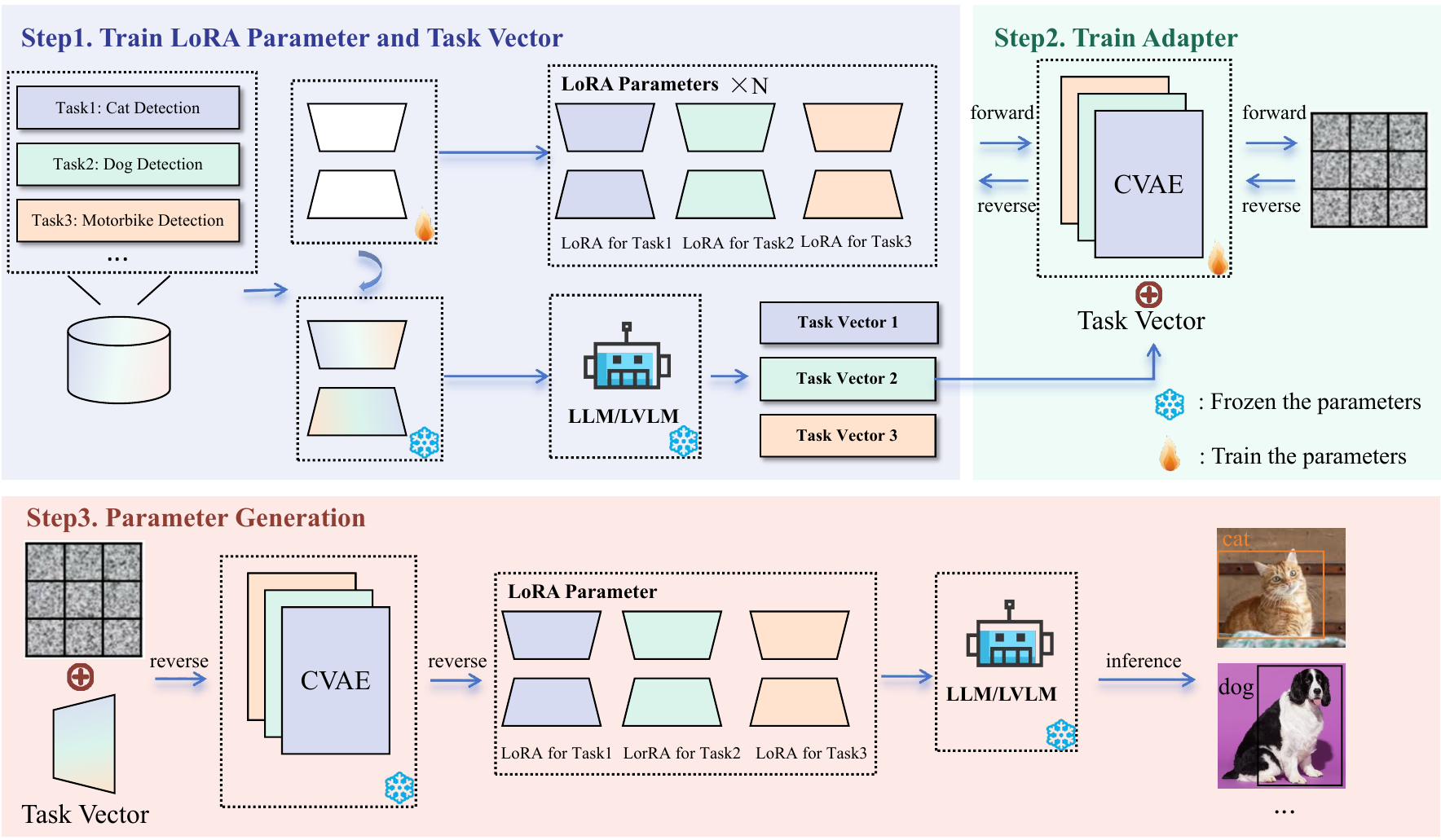}
    \caption{\textbf{Overall Framework of ICM-LoRA. }We first train different task LoRAs based on the dataset task categories and generate their task vectors. Then we train self-designed CVAE by utilizing these task data by in-context meta learning. Finally we achieve model customization by using text bootstrapping to make the trained CVAE generate the task LoRA we expect.}
    \label{fig:pipeline}
    \vspace{-0.4cm}
\end{figure*}
\section{Methodology}
In this section, we present our approach in terms of overview, task vector extraction , and parameter sampling and reconstruction for model customization.
\subsection{Overview}
As shown in Figure \ref{fig:sne}, we extracted the final hidden states of five different categories, dog, sofa, cat, bicycle and motorbike, at the last time step from the last layer of Florence-2's \cite{xiao2024florence} decoder and visualized them with S-NE \cite{van2008visualizing}. The visualization demonstrates that the hidden states from different categories form distinct clusters. For simplicity, we refer to the final hidden states at the last time-step from the last layer of decoders as 'task vectors' \cite{hendel2023context}.  

From the information mentioned above, we can draw two conclusions: first, task vectors from different categories are discriminative, as they form distinct clusters. Second, task vectors can represent the high-level features of different categories, given that the last layer and the final time step typically capture high-level representations.  

Therefore, task vectors satisfy the two key properties of condition vectors: discriminability and representativeness. Based on this, we hypothesize that task vectors can serve as condition vectors to effectively guide the generation process in the CVAE model. 

As shown in Figure \ref{fig:pipeline}, we provide an illustrative overview of the proposed method. Our method consists of three parts.
\begin{itemize}
  \item [1)] 
   \textbf{Preparing LoRA parameter data and extracting task vectors. }We fine-tune a Large Language Model (LLM) or a Large Vision-Language Model (LVLM) on a specific task category and save checkpoints from the final stages of the training process. These checkpoints serve as training data for the subsequent generative model. Next, we perform inference using the fine-tuned model on randomly selected samples from a specific task category, such as cat detection or dog detection. During inference, we extract the hidden states from the last layer at the final time step. These hidden states are then averaged to derive a task vector representing the specific task category.
  \item [2)]
  \textbf{Training the CVAE model. }The LoRA parameters extracted from the fine-tuned model checkpoints, along with the task vectors, are used as training data to train a Conditional Variational Autoencoder (CVAE). We use in-context meta learning to implement CVAE for modeling relationships between multiple tasks and to achieve enhancement of LoRA parameter distribution learning.
    \item [3)] 
  \textbf{Generating and applying LoRA parameters. }Using the trained CVAE, we sample from a Gaussian distribution to reconstruct the LoRA parameters for the target task category. The reconstructed LoRA parameters are then used to perform inference on the test set, enabling the model to generalize to the specific task effectively. 
\end{itemize}

\subsection{Task Vector Extraction}
Considering the contextual capabilities of large-scale models and evidence that in-context learning can generate task-specific representations \cite{hendel2023context}, we fine-tune a pre-trained model by LoRA \cite{hu2021lora} on a specific task category.  

Due to the hidden state corresponding to the last token, we extracted the hidden state of each sample $h _{i}^{\mathrm{last} } $ with the last token generated by LLM. For a specific set of task samples $\left \{ x_{i}  \right \} _{i=1}^{N} $, LLM generates hidden states of the task from the last layer.  These hidden states are then averaged to produce a compact \textit{task vector} which could be expressed as Eq. \eqref{Vt}:

\begin{equation}
{v} _{\mathrm{task} }=\frac{1}{N} \sum_{i=1}^{N} h_{i}^{\mathrm{last} } ,
\label{Vt}
\end{equation}
where $N$ is the number of task-specific samples, and ${v} _{\mathrm{task} } \in R^{d}  $ represents the task vector for the given category, and $d$ represented the dimensionality of the hidden state.

Since the last time-steps of many natural language processing scenarios contain complete information about the input sequence, we choose the last token to generate the hidden state $h_{i}^{\mathrm{last} } $. For the task of mapping an input sequence to a single vector, the hidden state of the last marker typically integrates information from all previous time steps, thus becoming a compact representation of the overall semantics of the input $x_i$.

Furthermore, we extract the hidden states from the final layer because this layer typically represents the most abstract and task-specific feature space. As information flows through the network, the first few layers typically encode general linguistic or structural features, while the last layer captures high-level semantic features specific to the current task. This recursion allows the final layer to act as a task feature extractor, providing a representation that is well suited for generating task vectors. The abstract nature of the last layer ensures that the generated task vector $x_{\mathrm{task} }$ can effectively capture the key features of the task.

\subsection{Conditional Variational Autoencoder}

Based on the Variational Autoencoder (VAE) \cite{kingma2013auto}, we employ a Conditional Variational Autoencoder (CVAE) to model the distribution of LoRA parameters conditioned on task vectors. The CVAE consists of an encoder $q_{\phi } \left ( z\mid l,v_{\mathrm{task} }  \right ) $, which maps the LoRA parameter $l$ and task vector $v_{\mathrm{task} } $ to a latent representation $z$, and a decoder $q_{\theta  } \left ( z\mid l,v_{\mathrm{task} }  \right ) $, which reconstructs $l$ from $z$ and $v_{\mathrm{task} } $. The encoder and decoder are conditioned on the task vector $v_{\mathrm{task} } $, which provides additional information to guide the generation of LoRA parameters.
% \begin{figure}[htbp] % h=here, t=top, b=bottom, p=page
%     \centering
%     \includegraphics[width=0.83\linewidth]{cvae.pdf}
%     \caption{\textbf{CVAE}}
%     \label{fig:cvae} 
% \end{figure}

The encoder $q_{\phi} \left ( z\mid l,v_{\mathrm{task} }  \right ) $ models the approximate posterior distribution over the latent variables $z$, given the LoRA parameter $l$ and the task vector $v_{\mathrm{task} } $. The encoder takes as input the concatenation of the LoRA parameter $l$ and the task vector $v_{\mathrm{task} } $, which is denoted as Eq. \eqref{eq1},
\begin{equation}
 x=\left [ l;v_{\mathrm{task} }  \right ] \in R^{d_{l} +d_{\mathrm{task} } },\label{eq1}
\end{equation}
where $d_l$ and $v_{\mathrm{task} } $ represent the dimensions of the LoRA parameter $l$ and the task vector $v_{\mathrm{task} } $, respectively. 

The concatenated vector $x$ is passed through a neural network, which outputs the parameters of the approximate posterior distribution. This process can be expressed as Eq. \eqref{q},
\begin{equation}
    q_{\phi } \left ( z\mid  l,v_{\mathrm{task}}\right )=\mathcal{N} \left ( z;\mu _{\phi },\sigma _{\phi }^{2}\left ( x \right )    \right ) ,
    \label{q}
\end{equation}
where $\mu _{\phi } \left ( x \right ) $ and $\sigma _{\phi }^{2} \left ( x \right ) $ are the mean and variance of the latent variable $z$, computed from the input $x$. 

The latent variable $z$ is then sampled from this distribution using the reparameterization trick, which could be represented as Eq. \eqref{z}:
\begin{equation}
    z=\mu _{\phi } \left ( x \right ) +\sigma _{\phi } \left ( x \right ) \odot \epsilon ,
    \label{z}
\end{equation}
where $\epsilon \sim \mathcal{N} \left ( 0,I \right ) $ is a noise term and $\odot $ represents element-wise multiplication.

The decoder $p_{\theta } \left ( l\mid z,v_{\mathrm{task} }  \right ) $ models the likelihood of the LoRA parameter $l$ given the latent variable $z$ and the task vector $v_{\text{task}}$. 
The decoder receives $z$ and $v_{\text{task}}$ as input, which are concatenated as Eq. \eqref{x'},
\begin{equation}
    x'=\left [ z;v_{\mathrm{task} }  \right ] \in R^{d_{z}+d_{\mathrm{task} }  } ,
    \label{x'}
\end{equation}
where $d_z$ is the dimensionality of the latent variable $z$. 

The concatenated vector $x'$ is then input into a neural network to output the parameters of the likelihood distribution for the LoRA parameter $l$, the process could be represented as Eq. \eqref{p_}:
\begin{equation}
    p_{\theta } \left ( z,v_{\mathrm{task} }  \right ) =\mathcal{N} \left ( l,\hat{\mu  } _{\theta }\left ( x' \right ), \hat{\sigma } _{\theta }^{2}\left ( x' \right )     \right ) ,
    \label{p_}
\end{equation}
where $\hat{\mu}_\theta(x')$ and $\hat{\sigma}_\theta^2(x')$ are the mean and variance of the predicted LoRA parameter $l$, computed from the input $x'$. 

The decoder aims to minimize the reconstruction error, ensuring that the generated LoRA parameters match the true parameters as closely as possible.
The latent space $z\in R^{k} $ is assumed to follow a Gaussian prior distribution which could be represented as Eq. \eqref{pz}:
\begin{equation}
    p\left ( z \right ) =\mathcal{N} \left ( z;0,I \right ) ,
    \label{pz}
\end{equation}
where $I$ is the identity matrix. 

\begin{table*}
\resizebox{\linewidth}{!}{ 
\begin{tabular}{c|cccccccccc}
\toprule 
               & \multicolumn{2}{c}{\textbf{Dog}} & \multicolumn{2}{c}{\textbf{Bicycle}} & \multicolumn{2}{c}{\textbf{Cat}} & \multicolumn{2}{c}{\textbf{Sofa}} & \multicolumn{2}{c}{\textbf{Motorbike}} \\
             \textbf{Method}        & \textbf{MAP50}  & \textbf{MAP75} & \textbf{MAP50}    & \textbf{MAP75}   & \textbf{MAP50}  & \textbf{MAP75} & \textbf{MAP50}  & \textbf{MAP75}  & \textbf{MAP50}     & \textbf{MAP75}    \\ 
             \midrule
Original LoRA              &     0.96           &        0.89        &      0.90             &         0.81        &  0.94               &        0.90        &        0.95         &        0.86         &          0.82          &  0.78                 \\
\midrule
Original Model         &     0.92           &      0.87          &         0.89          &        0.80          &    0.93             &        0.89        &       0.00          &          0.00       &           0.00         &        0.00            \\
Model Soup        &      0.93           &      0.87          &          0.90         &      0.78            &   0.93              &        0.88        &         0.81        &          0.74       &           0.81         &         0.72         \\

COND P-DIFF       &        0.94        &        0.87        &           0.90        &         0.77         &    0.93             &            0.89    &            0.84     &           0.78      &            0.80        &       0.74            \\

\textbf{ICM-LoRA} &        \textbf{0.96}          &     \textbf{0.89}           &      \textbf{0.90}             &         \textbf{0.81}         &  \textbf{0.95}               &        \textbf{0.91}        &        \textbf{0.95}         &        \textbf{0.86}         &           \textbf{0.83}         &  \textbf{0.78}                 \\
\bottomrule

\end{tabular}
}
    \caption{\textbf{Parameter Reconstruction Results for LoRA Rank $r = 2$ in Object Detection Task. }ICM-LoRA generates LoRA weights are closest to the original LoRA, and even better than the original LoRA in some tasks.}
    \label{tab-f} 
\end{table*}

The objective is to maximize the evidence lower bound (ELBO), which consists of two terms: the reconstruction term and the regularization term. The ELBO could be expressed as Eq. \eqref{Loss},

\begin{equation}
\begin{aligned}
\mathcal{L} = E_{q_\phi(z \mid l, v_{\text{task}})} 
&\left[ \log p_\theta(l \mid z, v_{\text{task}}) \right] + \\
& KL\left(q_\phi(z \mid l, v_{\text{task}}) \parallel p(z)\right),
\label{Loss}
\end{aligned}
\end{equation}
where $KL\left ( \cdot \parallel \cdot  \right ) $ represents the Kullback-Leibler divergence. The first term encourages the decoder to reconstruct accurate LoRA parameters, while the second term regularizes the latent space to match the Gaussian prior.

By conditioning both the encoder and decoder on the task vector $v_{\mathrm{task} } $, the model learns to generate LoRA parameters that are specific to the given task, leading to task-aware representations in the latent space.

In order for CVAE to generate task LoRA parameters, we utilize CLIP's \cite{radford2021learning} text encoder to output the task vector $v_{\mathrm{task} } $. A sample $z$ is drawn from the prior distribution $p(z)$, and the decoder generates the corresponding LoRA parameter:

\begin{equation}
l_{\text{generated}} = p_\theta(l \mid z, v_{\text{task}}).
\end{equation}

\section{Experiments}
In this section, we evaluated several tasks on LLMs and MLMs. We evaluated the task performance of the LoRA \cite{hu2021lora} generated by the current LoRA parameters generation methods. This demonstrates the effectiveness and reasonableness of our approach.

\subsection{Experiment Setting}

\textbf{Baselines.}
We chose the original model, LoRA, LoRA generated by Model Soup \cite{wortsman2022model} and COND P-DIFF \cite{jin2024conditional} as baseline to compare with our method and test the advantages of our method on different tasks.

\noindent \textbf{Datasets. }
For the computer vision task, we select the most representative target detection task as to conduct the experiment. We choose the COCO \cite{lin2014microsoft} dataset and divide it into different subclasses based on the detection task labels. For the language modelling task, we employ The Pile \cite{gao2020pile} as the training corpus. To simulate the multi-category training tasks, we pick five various subsets from the Pile corpus and validate our method on the test sets.

\noindent\textbf{Data Preparation.}
The model fine-tuning process also produces a series of LoRA matrices $\left \{ L_{t}  \right \} _{t=1}^{T} $ with different ranks $r$, where $T$ represents the number of fine-tuning steps. Each matrix $ L_{t}\in  \textbf{R}^{m\times n} $ is flattened into a one-dimensional vector $ l_{t}\in  \textbf{R}^{m\times n} $ to facilitate alignment with the task vector $v_{\mathrm{task} } $. These flattened LoRA parameters, along with the corresponding task vectors, form the training dataset $\left \{ \left ( v_{\mathrm{task}  } ,l_{t} \right )  \right \} $ for the self-designed CVAE.

\noindent \textbf{Training Strategies. }The CVAE model employs a 12-layer 1D CNN architecture for both the encoder and decoder. The loss function for the CVAE combines the Kullback-Leibler divergence (KLD) and reconstruction loss, with the KLD weight set to 0.005. The loss function could be expressed as Eq. \eqref{Loss} We fine-tuned the model on a specific task using LoRA (Low-Rank Adaptation) for a total of 150 epochs, saving the LoRA parameters from the final 50 epochs. The task vector is extracted from the last token of the last layer in the CLIP \cite{radford2021learning}. Subsequently, the CVAE model is trained for 2,000 epochs to ensure robust learning of the latent space. All experiments were conducted on a single NVIDIA A800 GPU, with each experiment taking approximately 3 hours to complete.

\noindent \textbf{Evaluation Metrics.}
We choose to record different LoRA ranks $r$ and parameter numbers $P$ in all experiment. For the object detection task, we chose to record Mean Average Precision (\textbf{MAP $\uparrow$}) at thresholds \textbf{IoU $=0.5,0.75$}. For the language modeling task, we chose perplexity (\textbf{PPL $\downarrow$}) and bits-per-character (\textbf{BPC $\downarrow$}) as the evaluation metrics.
\begin{table*}
\centering
\begin{tabular}{c|cccccccccc}
\toprule 
& \multicolumn{2}{c}{\textbf{ArXiv}} & \multicolumn{2}{c}{\textbf{Books}} & \multicolumn{2}{c}{\textbf{Ubuntu}} & \multicolumn{2}{c}{\textbf{Wikipedia}} & \multicolumn{2}{c}{\textbf{Gutenberg}} \\
\textbf{Method} & \textbf{PPL} & \textbf{BPC} & \textbf{PPL} & \textbf{BPC} & \textbf{PPL} & \textbf{BPC} & \textbf{PPL} & \textbf{BPC} & \textbf{PPL} & \textbf{BPC} \\ 
\midrule

Original LoRA & 6.75 & 0.41 & 7.07 & 0.48 & 9.66 & 0.58 & 5.54 & 0.43 & 8.60 & 0.62 \\ 
\midrule
Original Model& 7.76 & 0.44 & 7.67 & 0.51 & 10.00 & 0.59 & 6.07 & 0.46 & 8.75 & 0.63 \\
Model Soup &7.00 & 0.43 & 7.08 & 0.48 & 9.67&0.59 &5.56&0.45 &8.61 & 0.63 \\

COND P-DIFF &6.73 &0.41 & 7.10&0.49 & 9.67&0.58 & 5.55&0.44& 8.60&0.62 \\

\textbf{ICM-LoRA} &\textbf{6.74} &\textbf{0.40} & \textbf{7.07}&\textbf{0.48} & \textbf{9.65 }& \textbf{0.58 }& \textbf{5.54} & \textbf{0.43} & \textbf{8.59} & \textbf{0.61} \\
\bottomrule
\end{tabular}

    \caption{\textbf{Parameter Reconstruction Results for LoRA Rank $r = 2$ in Language Modeling.} Compared with baseline methods, ICM-LoRA generates LoRA parameters with a lower PPL ($\downarrow$) and BPC ($\downarrow$) on different subsets, equaling or even surpassing the original LoRA.}
    \label{tab-LM}
\end{table*}
\subsection{Main Results}

We conduct experiments on computer vision tasks and natural language processing tasks respectively, and demonstrate that our approach generalizes across models and can be adapted to tasks of multiple modalities.

\paragraph{Object Detection.}

% \usepackage{multirow}
% \usepackage{multirow}

\begin{table}
\centering
\begin{tabular}{c|cccc}
\toprule 
              & \multicolumn{4}{c}{\textbf{Rank}}                                                         \\
\textbf{Method}        & $r=1$                  &$ r=2  $                & $ r=4$                 &$ r=8 $                 \\ 
\midrule

Original COCO & \multicolumn{4}{c}{25G}                                                                   \\ 
\midrule
Original LoRA & 2.1G & 4.3G & 8.5G & 16.9G \\
Model Soup    &       423MB               &        453MB              &      478MB                &              504MB        \\

COND P-DIFF   &        314MB              &       314MB                &  318MB                     &              326MB         \\

\textbf{ICM-LoRA}      &       \textbf{283MB}                &       \textbf{283MB}                & \textbf{283MB}                      &        \textbf{283MB}               \\
\bottomrule
\end{tabular}
\caption{\textbf{Storage Memory Required for Different Methods. }In the vision task, ICM-LoRA and COND P-DIFF take up less storage overhead compared to the original dataset with the original LoRA parameter weights.}
    \label{tab-m} 
\end{table}

As shown in Table \ref{tab-f}, we selected several subsets of the more conventional tasks and fine-tuned on Florence-2 \cite{xiao2024florence}. the LoRA parameters generated by ICM-LoRA in the subset of expert tasks in the COCO dataset have the smallest difference in effect from the original LoRA parameters, and even the LoRA parameters generated by ICM-LoRA outperform the original LoRA in some of the tasks. this suggests that our method generates more LoRA parameters than the other methods. complete. By adding in-context learning, ICM-LoRA's understanding of task scenarios is enhanced compared to COND P-DIFF, implementing the effect of outperforming the original LoRA on some tasks.

As shown in Table \ref{tab-m}, our method significantly achieves task-specific dataset compression using much less storage than the original LoRA weights with the original dataset. The compression of the visual dataset is achieved through the parameter generation method, which significantly reduces the storage cost. This shows that our approach not only generates task-corresponding LoRAs more accurately, but also enables task-based data compression.

\begin{figure}[t]
    \centering
    \includegraphics[width=1\linewidth]{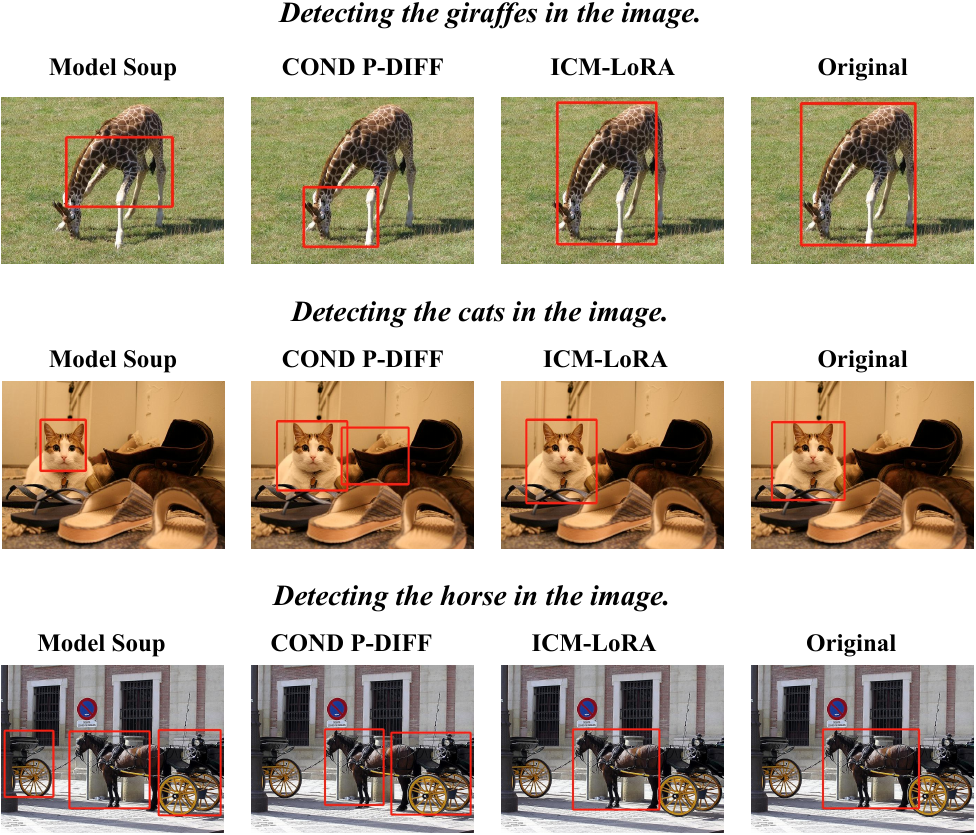}
    \caption{\textbf{Visual Comparison of Different Methods for Generating LoRA. }The LoRA generated by ICM-LoRA is most similar to the original LoRA effect.}
    \label{fig:vis}
\end{figure}

As shown in Figure \ref{fig:vis}, both Model Soup and COND P-DIFF were poorly labeled in the detection, and COND P-DIFF even had a false detection in the complex environment. These phenomena indicate that the reconstructed LoRA and original LoRA parameters of Model Soup and COND P-DIFF are so different that they cannot be successfully adapted to LVLM.

\paragraph{Language Modeling.}
When it comes to the language modelling task, we set the LoRA rank $r=2$. We fine-tuned the Llama-3-8B \cite{dubey2024llama} model across different tasks and proposed the result of five subsets from the Pile corpus, including ArXiv, Books, Ubuntu, Wikipedia, and Gutenberg. As shown in Table \ref{tab-LM}, compared to other methods, ICM-LoRA achieves the lowest perplexity and bits-per-character. This outperformance clearly shows its superiority in these specific tasks of language modelling.

In some subtasks, ICM-LoRA achieved the same PPL and BPC as the original LoRA and even achieve lower PPL and BPC than the original LoRA. This indicates that even in language tasks, ICM-LoRA can reconstruct the LoRA parameters and even reconstruct LoRA with more reasonable parameter distribution in some specific subtasks.

Since Llama-3 \cite{dubey2024llama} and Florence-2 \cite{xiao2024florence} have different parameter distributions and model constructions, the LoRA construction has a different parameter distribution. We argue that ICM-LoRA can adapt both multi-modal and language tasks, and can adapt different models with diverse parameter distributions. Therefore, we conclude that ICM-LoRA is highly effective in enhancing the performance of language modeling for parameter generation.

\subsection{Ablation Studies}
In this section, we will first discuss the effect of different ranks and parametric quantities of LoRA on the generation of task LoRA. Then we discuss the effect of different number of convolutional layers $n$ on model performance during sampling. Finally, we discuss the impact on different task vector generated by CLIP's vision encoder and text encoder.

\noindent \textbf{Impact of LoRA Rank $r$ and Parameter Number $P$.}
We trained the LoRA parameters with rank $r$ of 1,2,4,8 respectively and trained CVAE using these LoRA parameters. The task LoRA parameters were generated based on the task vectors and evaluated on the COCO dataset. Meanwhile, we tested several other methods and discussed the effect of LoRA parameter size on the generated LoRA parameters. We selected dogs and cats as examples and reported their \textbf{MAP50}.

\begin{table}[t]
\resizebox{\linewidth}{!}{ 
\begin{tabular}{cc|cccccc}
\toprule 
              & \textbf{}  & \multicolumn{2}{c}{\textbf{Model Soup}} & \multicolumn{2}{c}{\textbf{COND P-DIFF}} & \multicolumn{2}{c}{\textbf{ICM-LoRA}} \\
\textbf{Rank} & \textbf{$P$} & \textbf{Dog}       & \textbf{Cat}       & \textbf{Dog}        & \textbf{Cat}       & \textbf{Dog}      & \textbf{Cat}      \\ 
\midrule
r = 1           & 241241     &      0.93              &            0.93        &           0.94          &    0.93             &           0.95         &  0.97                 \\
r = 2           & 482482     &      0.93              &            0.93        &           0.94          &    0.93                &          0.96         &    0.95               \\
r = 4           & 964964     &     0.93               &            0.92        &            0.93         &        0.93            &            0.95       &     0.96              \\
r = 8           & 1929928    &          0.91          &         0.91           &           0.90          &        0.90            &          0.95         &   0.96                 \\ 
\bottomrule
\end{tabular}}
\caption{\textbf{Impact of LoRA Rank and Parameter Number. }Our method is more robust in higher rank and more parameter LoRA compared to baselines.}
    \label{tab-a1} 
\end{table}
\begin{figure}[t]
    \centering
    \includegraphics[width=1\linewidth]{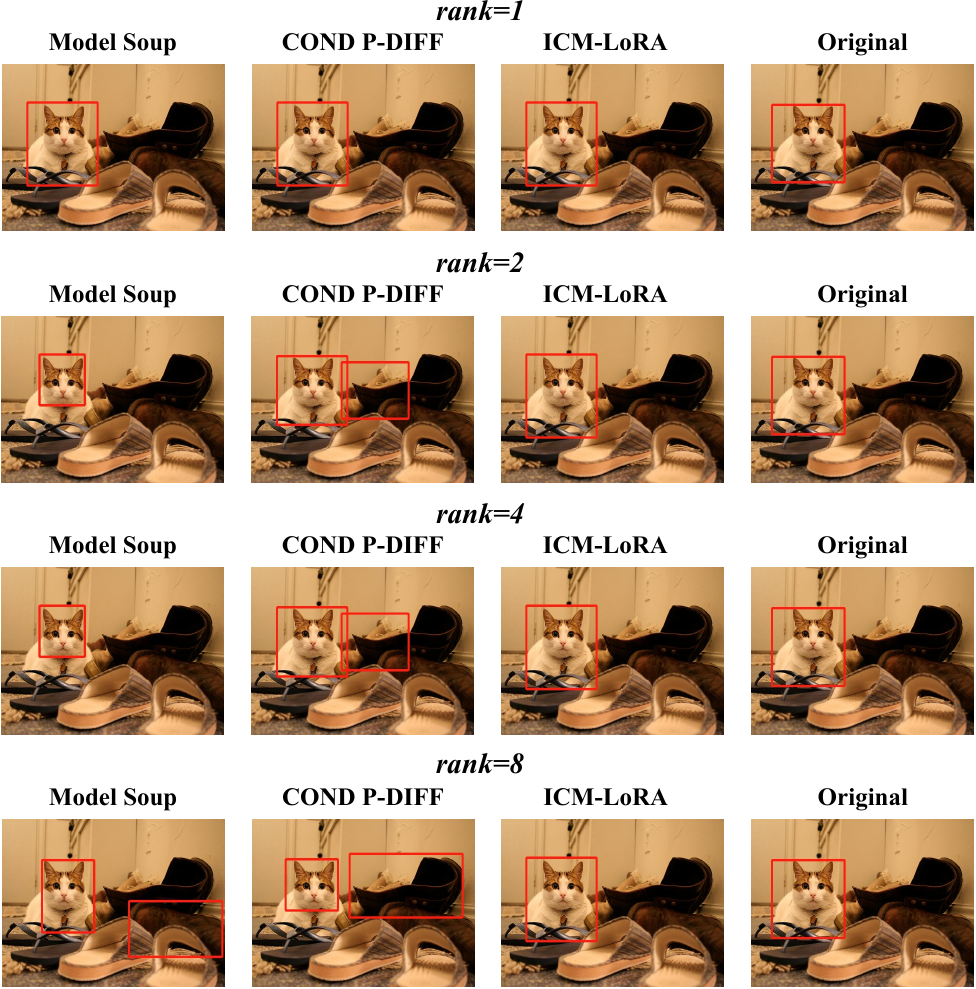}
    \caption{\textbf{Visualization of LoRA Rank Impacts. }For the task \textit{``Detecting cats in the image''.}, ICM-LoRA is less affected by LoRA rank compared to other methods.}
    \label{fig:vis2}
\end{figure}

As shown in Table \ref{tab-a1} and Figure \ref{fig:vis2}, as the rank $r$ of LoRA gradually increases, the number of LoRA parameters also increases. The other methods gradually decrease the effect on LoRA reconstruction as the number of LoRA parameters increases, which indicates that these methods cannot adapt to the reconstruction of LoRA with a large number of parameters. The detection effect of our method is almost the same as the original LoRA with the increase of LoRA parameters, which indicates that our method is more robust and can adapt to the reconstruction of LoRA weights with different numbers of parameters.

\noindent \textbf{Impact of Convolutional Layers Number.}
We evaluate the effect of sampling convolutional layers with different LoRA ranks and number of layers on the model's generation of LoRA weights for the task \textit{“Detecting Cats in Pictures”}.
\begin{table}[t]
\centering
\begin{tabular}{c|cccc}
\toprule 
     & $r=1$ & $r=2$ & $r=4$ & $r=8$ \\ 
     \midrule
$n=10$  & 0.93    &  0.93   &  0.91   &  0.89   \\
$n=11$  & 0.93    &  0.93   &   0.92  &   0.91  \\
$n=12$ &  0.93   &  0.93   &   0.93  & 0.93    \\
$n=13$ &  0.93   &  0.93   &   0.93  & 0.93    \\
$n=14$ &  0.93   &  0.93   &   0.93  & 0.93    \\
$n=20$ &  0.90   &  0.91   &   0.92  & 0.93    \\ 
\bottomrule
\end{tabular}
\caption{\textbf{Impact of Sampling Convolutional Layers. }LoRAs with larger ranks require more convolutional layers to be sampled, but too many convolutional layers can lead to poor model sampling.}
    \label{tab-a2} 
\end{table}

As shown in Table \ref{tab-a2}, as the LoRA rank and the number of parameters rise, the effect of sampling progressively decreases the fewer the convolutional layers. This indicates that the deeper the network is, the better the model samples the parameters. However, when the network convolution layer is too deep, it causes the model to fail to learn the parameter distribution characteristics of the fewer parameter LoRA. Therefore, we choose to use 12 convolutional layers to sample the LoRA parameters in our experiments. 

\noindent \textbf{Impact on Text and Vision Task Vector.} 
We generate text task vectors using \textit{“Detect the cat in the picture.”} and \textit{“Detect the dog in the picture.”} by text encoder. Then generate vision task vectors using cat and dog images. Finally we evaluate the reconstructed LoRA parameters on a subset of cats and dog in COCO and report \textbf{MAP50}.
\begin{table}[t]
\centering
\begin{tabular}{cc|cc}
\toprule 
\textbf{Vision} & \textbf{Text} & \textbf{Dog}  & \textbf{Cat}  \\ \midrule
   √   &   ×  & 0.96 & 0.95 \\
×      & √    & 0.96 & 0.95 \\
   √  &   √  & 0.96 & 0.95 \\ 
   \bottomrule
\end{tabular}
\caption{\textbf{Impact of Text and Vision Task Vector. }The visual task vectors and the text task vectors reconstructed by LoRA have essentially the same effect in both tasks.}
    \label{tab-a3} 
\end{table}

As it shown in Table \ref{tab-a3}, task vectors instructed by vision and task input has equal impact in generate the task vector， So we consider that task vectors of different modalities have equivalent effects on parameter generation. And the simultaneous use of multimodal task vectors is not possible for the process of enhancing parameter generation.
\section{Conclusion}
In this paper, we propose ICM-LoRA, a novel framework that uses a self-designed parameters generator, Conditional Variational Autoencoder (CVAE), which could generate LoRA parameters to implement model customization. ICM-LoRA achieves task vectors and LoRA parameter context modeling by combining in-context learning and meta-learning, which allowing CVAE to learn LoRA parameter distributions more accurately.

Our method achieves accurate task instruct LoRA parameter generation with only CVAE, eliminating the need for additional training data and storage. The experimental results on both language modeling and object detection tasks have further validated the effectiveness of our approach could apply to different models with different tasks. ICM-LoRA can also reduce storage costs and improve computational efficiency. Overall, ICM-LoRA represents a significant advancement in parameter generation and large-scale model customization.

\clearpage
\bibliographystyle{named}
\bibliography{ijcai25}
\end{CJK}
\end{document}